\documentclass[11pt]{article}
\usepackage[utf8]{inputenc}
\usepackage[margin=1in]{geometry}
\usepackage{amsmath,amssymb}
\usepackage{graphicx}
\usepackage{booktabs}
\usepackage{array}
\usepackage{tabularx}
\usepackage{multirow}
\usepackage{xurl}
\usepackage[numbers,sort&compress]{natbib}
\usepackage{authblk}

\usepackage[normalem]{ulem}

\usepackage[utf8]{inputenc}
\usepackage{graphicx,epsfig}
\usepackage{epstopdf}
\usepackage{amssymb,amsmath,amsthm,amsfonts}
\usepackage{bm}
\usepackage[dvipsnames]{xcolor}
\usepackage[linktocpage,breaklinks]{hyperref}
\hypersetup{colorlinks=true,
            citecolor=NavyBlue,
            linkcolor=NavyBlue,
            urlcolor=NavyBlue}

\title{Rater State Bias in RLHF Preference Data: An Audit Framework}

\author[1]{Elena Kopteva\thanks{\texttt{koptieva@illinois.edu}}}
\affil[1]{The Grainger College of Engineering,
Department of Physics \& Illinois Center for Advanced Studies of the Universe, University of Illinois Urbana-Champaign, Urbana, Illinois 61801, USA}

\author[2,3]{Vitaliy Hlynianyi-Zhuk\thanks{\texttt{hlyniany@gmail.com}}}
\affil[2]{Faculty of Applied Mathematics,
Oles Honchar Dnipro National University, Dnipro, 49045, Ukraine}
\affil[3]{Department of Clinical Psychology,
Kyiv Institute of Modern Psychology and Psychotherapy, Kyiv, 01133, Ukraine}

\date{\today}

\begin{document}

\maketitle

\begin{abstract}
We identify a structured confound in Reinforcement Learning from Human Feedback (RLHF). Pairwise preference labels are intended to reflect the compared outputs, but they may also reflect the rater's state during annotation. Under sustained stressful or distressing conditions, raters' preferences may shift over time, so that preference data encode rater state alongside judgments about response quality. We argue that, if present, such shifts would differ from ordinary disagreement or random label noise. They would be state dependent, could be shared across annotators under similar conditions, and would not necessarily cancel during aggregation, reward modeling, and policy optimization. We propose rater state shift as a plausible and testable source of structured bias in RLHF preference data.

This paper develops a hypothesis and an audit framework for studying this source of bias. We define rater state shift, rater state confound, and correlated rater state bias. We also propose survival level emotional authenticity as a candidate output signature, defined by lexical, pragmatic, discourse, and safety features whose reliability and validity remain to be demonstrated. We show that systematic rater state bias can survive aggregation and may enter the learned reward signal. We state five testable predictions, together with effect size thresholds for an initial audit, and note which require proprietary data. Finally, we present an audit protocol and pilot study plan that can be applied to publicly available instruction tuned models. We do not infer the training history of any specific deployed model.
\end{abstract}
\newpage

\tableofcontents


\section{Introduction}

Reinforcement Learning from Human Feedback (RLHF) is widely used to align large language models with human preferences \citep{christiano2017deep, ouyang2022training}. We study the preference data used in RLHF as a potential site of structured annotation bias. The core object is a pairwise preference label: given two model outputs, a rater records which one is preferred, and these labels become targets for reward modeling. Therefore, a preference label is not a direct measurement of output quality, it is a situated record of comparative judgment.

A growing body of work shows that annotator identity, beliefs, and cultural background shape labeled data \citep{sap2022annotators, santy2023nlpositionality, gordon2022jury}. This has motivated approaches that treat disagreement as structured signal rather than noise \citep{aroyo2015truth, plank2022problem, davani2022dealing}. Here, we focus on a different source of structure: the rater's state during annotation, rather than relatively stable rater traits. We ask how annotation conditions can systematically alter what raters reward. Under annotation conditions, we mean the external circumstances under which the work is performed. Under rater state, we mean the rater's current internal configuration under those circumstances, including affect, attention, and regulation.

We introduce the following definitions: a \textit{rater state shift} is a systematic change in a rater's state during annotation, shaped by annotation conditions and broader surrounding pressures, and sustained over time; a \textit{rater state confound} is a case where the recorded preference label reflects that state, rather than a stable judgment about output quality. A rater state shift becomes a rater state confound when it enters the recorded preference label. When such confounds systematically shift aggregate preferences, they can bias the learned reward signal even if annotations are independent. If the confounds are also correlated across raters, we refer to the resulting case as \emph{correlated rater state bias}.
We do not claim to identify the training history of any specific deployed model. Our claim is that rater state bias is a plausible and testable source of distortion in RLHF preference data.

In practice, RLHF preference labels are generated by paid annotators working inside managed data production systems, often through crowdwork or contractor arrangements rather than under idealized laboratory conditions \citep{gray2019ghost, roberts2019behind}. In NLP and dataset creation more broadly, annotation quality depends not only on the item being labeled, but also on workforce selection, training, qualification filters, feedback, compensation, and task design \citep{huang2023worker, klie2024analyzing}. 
These operational conditions matter here because they provide a natural pathway by which annotation can become state dependent: what raters notice, tolerate, reward, or reject may shift with the conditions under which labeling is performed.

We propose a concrete mechanism linking introduced definitions to the training workflow. Rater state confounds can produce systematic shifts in aggregate preferences even when annotations
are independent (Section~\ref{sec:entry}). Shared annotation conditions may also induce correlation across raters, reducing the effective sample size.

If reward model training absorbs the shifted preferences, the learned reward function can reflect this bias (Section~\ref{sec:absorption}). During  policy optimization, the KL-constrained policy then optimizes against whatever the reward model has learned, including the absorbed confound, which can amplify the distortion in generated outputs (Section~\ref{sec:amplification}).

This paper presents a hypothesis and an audit framework. It does not argue that any observed model behavior can be reduced to a single cause. The central point is that if annotation conditions can systematically affect raters during preference labeling, then RLHF preference data may contain a structured and testable source of distortion that should be examined directly.

The paper is organized as follows. 
Section~\ref{sec:related} reviews related work on annotator bias, signal imperfection in RLHF, and measuring response style in text.
Section~\ref{sec:rlhf} formalizes preference training as preference signal estimation and introduces the rater state framing. 
Section~\ref{sec:scaling} traces how rater state bias propagates through RLHF. 
Section~\ref{sec:evidence} summarizes the empirical premise for rater state shift. 
Section~\ref{sec:hypothesis} states the hypothesis and falsifiable predictions.
Section~\ref{sec:measures} defines the measurement taxonomy, audit protocol, and pilot study. 
Section~\ref{sec:discussion} discusses implications and limitations, and Section~\ref{sec:conclusion} concludes.


\section{Related Work}
\label{sec:related}
This section brings together several lines of related work that frame the phenomenon we study. In this paper, we consider the possibility that RLHF preference labels may be systematically distorted by changes in the rater's state during annotation, and that such distortions may persist through reward modeling and downstream training. The studies reviewed here are relevant because they show that labeled data often reflect annotators as well as items, that imperfections in learned reward signals can be amplified during optimization, and that response style can be audited with reproducible measures at the level of the text.

The closest prior work is the literature on annotator bias, disagreement, and positionality. That literature shows that label variation often has structure rather than being mere noise. Our contribution is to extend that logic from relatively stable annotator characteristics to within rater variation induced by annotation conditions. We locate rater state shift, rater state confound, and correlated rater state bias in that space.


\subsection{Sycophancy and Preference-Driven Bias in RLHF}

The most directly related line of work concerns sycophancy: the tendency of RLHF-trained models to favor responses that affirm a user's stated or implied beliefs over more truthful or better-judged ones. \citet{perez2022discovering} documented sycophantic behaviors through model-written evaluations and observed that they tend to increase with model scale. \citet{sharma2023towards} demonstrated sycophancy across several production assistants and, by analyzing existing human preference data, found that responses matching a user's views are more likely to be preferred, indicating that human preference judgments themselves contribute to the behavior. Concurrent and independent formal work by \citet{shapira2026rlhf} analyzes how alignment from human feedback amplifies sycophancy: working within the Bradley-Terry framework for reward learning from pairwise comparisons, they characterize when bias in annotators' preferences induces a reward gap, and show that the direction of behavioral drift is governed by a covariance, under the base policy, between an endorsement signal and the learned reward, with a first-order effect reducing to a mean-gap condition.
Our work shares the premise that annotator preference bias can be absorbed into the reward model and amplified during policy optimization, and the Bradley-Terry amplification logic we sketch in Section~\ref{sec:scaling} is consistent with the mechanism that \citet{shapira2026rlhf} analyze formally. We differ in the \emph{source} of bias we examine. Sycophancy concerns a content relation between prompt and response, namely agreement with a user's expressed belief. We instead consider rater \emph{state} during annotation: a shift in the rater's affective and regulatory condition under sustained operational strain, which may alter what is rewarded independently of whether the response endorses any user belief. We further emphasize a structural property that the sycophancy literature does not foreground: because rater state is shaped by shared annotation conditions, the resulting bias may be \emph{correlated across raters}, so that standard agreement-based quality control does not remove it. Rater state shift can thus be viewed as a distinct instance of preference-driven bias, with a different trigger and a correlation structure that our audit framework is designed to detect.


\subsection{Annotator Bias, Disagreement, and Positionality}

A considerable body of research has shown that labeled data are shaped not only by the items being annotated, but also by the people who annotate them. \citet{sap2022annotators} showed that annotator beliefs and identities influence toxic language labels. \citet{gordon2022jury} introduced jury learning as a way to retain dissenting annotator perspectives rather than collapsing them into a single majority label. \citet{santy2023nlpositionality} showed that NLP datasets and models disproportionately align with Western, white, college educated, and younger populations. \citet{plank2022problem} argued that human label variation should often be treated as informative signal rather than noise, and \citet{davani2022dealing} showed that models preserving individual annotator perspectives can outperform majority vote aggregation on subjective tasks. More broadly, work on intersectional fairness in machine learning has emphasized that socially structured differences do not reduce to a single demographic axis and can be obscured when annotation and evaluation processes treat human judgments as uniform \citep{GoharCheng2023IntersectionalFairness}.

This literature is the closest point of contact for our argument. It shows that disagreement in labeled data often has structure, and that this structure can reflect perspective, position, and social experience rather than annotation error alone. In most of this work, however, the relevant sources of variation are treated as relatively stable across time. They are tied to who the annotator is, how the annotator is situated, or what background the annotator brings to the task.

Our contribution is to shift attention to a different source of structured variation: changes in the rater's state during annotation. The key idea is not that raters differ from one another in stable ways, but that the same rater may judge differently under different annotation conditions. This moves the discussion from persistent annotator characteristics to variation that can arise during the annotation process itself. In that sense, our framework extends the logic of structured disagreement into a setting where the relevant source of variation is not fixed perspective, but state dependent judgment under operational conditions.


\subsection{Signal Imperfection and Amplification in RLHF and Machine Learning}

A central concern in alignment is that optimization can amplify imperfections in the signal being optimized. In RLHF, \citet{gao2023scaling} characterized reward model overoptimization, showing that policy optimization against an imperfect learned reward can degrade true performance in a Goodhart-like way, through optimization of an imperfect proxy for the intended target. They also derived scaling laws for this effect. \citet{rafailov2024scaling} extended this concern to direct alignment methods, showing that related forms of overoptimization can arise even without an explicit reward model.

A broader machine learning literature shows a similar pattern. \citet{zhao2017men} demonstrated that image classifiers can amplify gender activity correlations beyond their rates in the training data. \citet{wang2021directional} formalized directional bias amplification and showed that it depends in part on the relative difficulty of recognizing group membership and class membership. \citet{hall2022systematic} provided the first systematic study of bias amplification, showing that the effect varies with model capacity, training set size, and training duration, and is not monotonic in dataset size. More recently, \citet{subramonian2024effective} showed analytically, in ridge regression, that overparameterization can amplify group level bias even without class imbalance, producing disparities that do not disappear simply by increasing model size.

Taken together, these studies support a general point that is important for our argument: imperfections in the training signal can survive learning and become stronger through optimization. Our contribution is to identify a particular source of such imperfection in RLHF preference data. If annotation conditions induce systematic rater state shifts, and if those shifts enter recorded preference labels as rater state confounds, then the learned reward signal may reflect more than the intended target of comparison. When those confounds are shared across raters and retained through training, they create a plausible path from annotation conditions to correlated rater state bias in downstream model behavior. Section~\ref{sec:scaling} develops this argument  in detail.


\subsection{Measuring Response Style and Discourse Structure in Text}

Computational methods for studying response style in text are well developed. Within this broader space, tools for measuring affect and emotion provide a particularly robust and reproducible set of measures. Lexical resources such as LIWC \citep{pennebaker2015liwc}, VADER \citep{hutto2014vader}, and the NRC Emotion Lexicon \citep{mohammad2013nrc} provide reproducible measures of sentiment and emotion categories. Classifiers trained on datasets such as GoEmotions \citep{demszky2020goemotions} and EmpatheticDialogues \citep{rashkin2019towards} extend this line of work to finer distinctions in emotion and empathic response. Together, these tools show that important aspects of response style can be measured in a systematic and replicable way.

We use this literature in a limited and practical sense. These tools do not reveal inner state, and we do not treat them as ground truth about what a speaker or model is feeling. Instead, we use them as reproducible indicators of observable properties of responses, including sentiment, emotional intensity, and cues associated with empathy or validation. They are useful here because they are well validated, easy to reproduce, and suitable for comparison across feature families and datasets.

At the same time, standard affective measures do not fully capture the construct we study. They can identify positive valence, emotional language, or empathic wording, but they are often insensitive to pragmatic structure. A measured therapeutic response and an immediately validating one may both receive positive sentiment scores, even though they differ in how quickly validation appears, how strong it is, and how much of the response is devoted to validation before redirection. For this reason, our measurement framework does not rely on sentiment alone. It combines affective indicators with discourse features that capture the organization of the response itself.

This limitation is well known in practice. Sarcasm can invert apparent sentiment while preserving surface wording. Politeness norms can soften negative stance or mask emotional intensity. Genre also matters: customer support, therapeutic dialogue, and informal chat can share a warm surface style while differing in function and discourse structure.

Work on supportive communication, attachment, and psychotherapy process also helps motivate several discourse features relevant to our construct. This literature discusses patterns such as early emotional acknowledgment, strong validation, affect mirroring, and the timing of advice relative to emotional contact \citep{bowlby1969attachment, mikulincer2003attachment, waters2019attachment, ainsworth1978patterns, rogers1957necessary, truax1967toward, stern1985interpersonal, feeney2004transfer}. We use this literature narrowly. We do not infer attachment style, diagnosis, or therapeutic quality from text. Rather, we draw on it to motivate observable discourse features that can be operationalized as measures and used to distinguish generic supportive tone from the more intense pattern later defined as survival level emotional authenticity. These considerations motivate the broader feature taxonomy introduced in Section~\ref{sec:measures}.


\section{RLHF as Preference Signal Estimation Under Rater State Bias}
\label{sec:rlhf}

This section separates four things that are often blended together in RLHF discussions:
(i) the standard estimator and its stable target assumption;
(ii) rater state shift as an alternative account of preference labels;
(iii) what RLHF preference data records and what it does not;
(iv) the observable signatures that would support or weaken the hypothesis.


\subsection{Technical Foundations}

We begin with the standard RLHF estimator and the observables typically available for analysis.

RLHF aligns language models with human preferences through three stages \citep{christiano2017deep, ouyang2022training}.
(1) Supervised Fine-Tuning (SFT), when a pretrained model is fine-tuned on curated prompt-response pairs.
(2) Reward Model Training, when human raters compare outputs per prompt, and these comparisons train a reward model that assigns a scalar preference score to each response, denoted $r_\theta(x,y)$ for response $y$ given prompt $x$.
(3) Policy Optimization, the language model is fine-tuned to produce responses that receive higher scores from the reward model $r_\theta$, while staying close to the SFT model via a KL penalty.

A common abstraction is that there exists a stable target preference
probability: for a given prompt $x$, one response $y$ has some fixed
probability of being preferred to another $y'$ under an idealized
preference process.
Rater disagreement is then treated as noise around a fixed latent
preference function.
If each pairwise label is a noisy sample of the same stable target preference
probability, with enough labels, aggregation methods (majority vote,
Bradley-Terry, or similar) approximate that target rather than
individual rater noise.

In RLHF datasets available for analysis, one typically observes prompts $x$, candidate responses $y,y'$, and a pairwise label indicating which response was preferred.
Timestamps are often available.
Rater IDs may be available, but not reliably.

Direct measures of rater state and exposure history are usually unavailable.
This includes prior content seen, time on task, breaks, workload, and support resources.
Detailed task routing logs are also often missing.


\subsection{Rater State Shift and the Stable Target Assumption}

We now introduce rater state shift as a source of bias in RLHF preference data.

The standard estimator assumes that pairwise labels are noisy samples from a single stable target preference process.
Our alternative is that judgments may shift with rater state during annotation.

To express this, suppose each judgment is produced under a latent rater state $s$.
Let $r(x,y \mid s)$ denote a latent preference score for response $y$ to prompt $x$ in state $s$.
According to the Bradley-Terry model \citep{bradley1952rank}, which underlies RLHF preference modeling \citep{christiano2017deep}, this induces the state-dependent pairwise preference probability
\begin{equation}
p_s(y \succ y' \mid x)
=
\sigma\!\bigl(r(x,y \mid s)-r(x,y' \mid s)\bigr),
\label{eq:state_bt}
\end{equation}
where $\sigma$ is the sigmoid function.

If judgments are produced under different rater states, then the pairwise preference signal entering reward model training is the average over states:
\begin{equation}
\bar p(y \succ y' \mid x)
=
\mathbb{E}_{s \sim S}\!\left[p_s(y \succ y' \mid x)\right],
\label{eq:mixture}
\end{equation}
\noindent where $S$ denotes the distribution of rater states in the data.

The learned reward model $r_\theta(x,y)$ is then fit to this mixed pairwise signal through its implied Bradley-Terry probability,
\begin{equation}
p_\theta(y \succ y' \mid x)
=
\sigma\!\bigl(r_\theta(x,y)-r_\theta(x,y')\bigr).
\label{eq:bt_mixed}
\end{equation}

If $S$ is tightly concentrated, this reduces to the standard stable target picture.
If shifted states are common in the data, then the learned reward model is fit to a mixture of state dependent preferences rather than to a single fixed preference process.

If rater state effects are independent across raters and over time and centered around a stable target preference, they act like additional label noise.
Averaging over many labels will then tend to attenuate their effect as the dataset grows.

This need not hold when the distribution of rater states is persistently shifted.
If some states occur more often over an extended period and systematically affect judgments in the same direction, their influence can survive aggregation even when individual rater states are independent.
Positive correlation across raters can additionally reduce the effective number of independent observations, limiting the precision gained through averaging.

This distinction guides the audit framework.
The next two subsections explain why structured rater state shifts are plausible and how they can be studied through observable proxies.


\subsection{Shared Conditions and Exposure Proxies}
\label{sec:tasks}

We next explain why such bias could be correlated across raters and how it might still be studied when direct measures of rater state are unavailable.

RLHF annotation includes different task types.
Some tasks collect preference rankings over model outputs such as helpfulness, harmlessness, or honesty.
Others involve safety or policy labeling for harmful content and policy violations.

Most public evidence about psychological strain among annotation workers comes from safety labeling and content moderation.
There is less direct public evidence about preference ranking, especially for supportive dialogue.
This matters because it limits what can be claimed from existing reports.

Two distinct pathways should be separated here, because the mechanism may operate through either one or through both.

First, there is a workforce carryover pathway.
Workers may rotate across safety labeling and preference ranking under the same vendors, incentives, and monitoring conditions \citep{perrigo2023exposed, gray2019ghost}.
If so, strain induced in one task type could influence judgments made in another.

Second, there is a direct exposure pathway within preference work itself.
Preference ranking can involve traumatic or distressing material when prompts concern self harm, abuse, violence, or crisis.
In that case, the relevant state shift need not be imported from safety labeling.
It can arise during preference annotation itself.

These two pathways imply different empirical signatures.
The carryover pathway points to task routing, assignment history, and lagged effects of prior exposure.
The direct exposure pathway points to within session drift and topic conditioned shifts as the recent share of sensitive prompts rises.
Public reporting also describes low pay, precarious contracts, high volumes, and limited support as broad features of data work rather than features of one narrow task \citep{cbsnews2024moderators, sfgate2024scaleai}.

Because direct measures of rater state and exposure are usually unavailable, the audit must rely on observable proxies.
Recent exposure can be approximated by the share of sensitive prompts in a rater's recent window, or in the broader batch during a given time interval.
Session structure can be approximated from timestamps, including session duration and gaps between judgments.
Topic clusters can be constructed from prompt embeddings or simple taxonomy tags when available.
Shift timing can be studied through change points in label distributions, especially around operational events such as rubric updates, model version changes, or vendor policy changes.

These proxies do not identify mechanism on their own.
They do let us test whether preference drift aligns with plausible sources of strain under real annotation conditions.


\subsection{Workforce Structure and Correlated Bias}
\label{sec:workforce}

Large scale annotation work does not imply statistical cancellation when judgments are produced under shared conditions.
What matters here is not only the number of labels, but also how annotation work is organized across vendors, regions, and time blocks.

Public reporting provides only partial information, but it shows a recurring pattern.
Large AI companies rely on third party vendors to staff labeling and evaluation work.
Reported setups include concentrated teams, low take home pay, and billing structures in which vendor rates greatly exceed worker pay \citep{perrigo2023exposed, cbsnews2024moderators}.
Other reporting describes annotation platforms operating at the scale of tens of thousands of workers \citep{sfgate2024scaleai}.

These facts support a narrower point.
Large scale annotation work is often organized through a small number of vendors and platforms.
Workforces may be regionally concentrated and exposed to shared policies, tooling, and schedules.
Under such conditions, rater state bias may be correlated across raters.

At the same time, public sources rarely permit clean identification.
They typically do not report (i) overlap between safety labelers and preference raters, (ii) task routing across assignments, or (iii) exposure histories over time.
Public evidence therefore does not show that exposure from safety labeling carried into preference ranking.
It does support treating correlated rater state bias as a plausible hypothesis rather than a proven mechanism.
This is why the audit must look for observable signatures in the preference data itself.

If correlated rater state bias is present, it should leave observable traces in preference data.
We would expect label distributions to drift with time on task and over calendar time.
We would also expect topic conditioned variation, because sensitive material is not uniformly distributed across prompts.

When rater IDs are available, we would expect within session drift and cross rater synchrony around shared operational changes, such as rubric updates, policy changes, or sudden shifts in sensitive content volume.
When rater IDs are unavailable, we would still expect time-based discontinuities and topic conditioned drift.

The hypothesis is weakened if label distributions remain stable over time after controlling for prompt topic, model version, and rubric changes.
It is also weakened if no within session drift appears when rater IDs are available, or if apparent drift disappears under simple operational covariates.
The hypothesis is strengthened if drift persists under these controls and aligns with the exposure proxies described above.


\section{How Rater State Bias Propagates Through RLHF}
\label{sec:scaling}

A natural objection to our hypothesis is that any rater state bias introduced by a subpopulation of distressed raters would be diluted to insignificance by the scale of RLHF. This section provides an analytical argument that, under stated modeling assumptions, 
systematic rater state bias can survive aggregation, enter the learned reward signal, and influence the response probabilities assigned by the optimized policy. To describe the resulting absorption-amplification mechanism we use the Bradley-Terry model, which is analyzed formally by \citet{shapira2026rlhf} for the case of belief-endorsement sycophancy.  

Our aim is to trace how systematic rater state bias
can propagate through RLHF and examine the role of
correlation arising from shared annotation conditions.
The parameter values below are illustrative and
require empirical validation. 

In what follows, rater state confound refers to the distortion at the level of recorded preference labels, while rater state bias refers to its aggregate retention in the learned preference signal.


\subsection{Entry into the Preference Signal}
\label{sec:entry}

We begin with a simple mean field model of rater state shift, which is a particular case of the setup in Section~\ref{sec:rlhf}.

Consider a preference annotation task in which each annotation compares a pair of model responses $(y,y')$ for a prompt $x$. Let $p_0(y \succ y' \mid x)$ denote the probability of preferring $y$ over $y'$ in the absence of rater state shift, and let $p_{\mathrm{shift}}(y \succ y' \mid x)$ denote the corresponding probability when rater state shift is present. For brevity, in the remainder of this section we omit explicit arguments in preference probabilities where this does not cause confusion. We write
\begin{equation}
p_{\mathrm{shift}} = p_0 + \delta(x),
\label{eq:preference_shift}
\end{equation}
where $\delta(x) \in \mathbb{R}$ represents the preference shift induced by rater state shift and may vary across prompt classes. For prompts where rater state does not affect the judgment, $\delta(x)$ should be near zero. For prompts where it shifts judgments in favor of the response pattern of interest, $\delta(x)$ is positive; when it shifts judgments away from that pattern, $\delta(x)$ is negative. 

Let $f$ denote the fraction of annotations generated under conditions where rater state shift is present. The aggregate preference probability in Eq.~\eqref{eq:mixture} observed by the reward model is then
\begin{equation}
\bar p(y \succ y' \mid x) = (1-f)\,p_0 + f\,p_{\mathrm{shift}} = p_0 + f\delta(x).
\label{eq:aggregate}
\end{equation}
Here, $f\delta(x)$ is the aggregate mean shift under the assumed mixture model.

A systematic shift remains in the aggregate preference signal even when annotations are independent. Positive correlation among annotations does not change the mean shift $f\delta(x)$ in this model. It reduces the effective sample size and the precision with which the shift can be estimated.

In practice, RLHF raters are often clustered: they may work for the same vendor, in the same region, under similar conditions, and with similar content exposure. To illustrate the effect of clustering, we use the Kish design effect for the idealized case of equal cluster sizes \citep{kish1965survey}. For this illustration, we assume one annotation per rater. Let $\rho$ denote the intraclass correlation coefficient within a cluster of size $n$, drawn from a total workforce of $N$ raters. The Kish design effect is
\begin{equation}
  D_{\mathrm{eff}} = 1 + (n-1)\rho,
  \label{eq:deff}
\end{equation}
and the effective number of independent observations is
$N_{\mathrm{eff}} = N / D_{\mathrm{eff}}$. For a cluster of $n = 200$
workers with $\rho = 0.1$, we get $D_{\mathrm{eff}} = 20.9$, so a
nominal workforce of $N = 1{,}000$ yields only
$N_{\mathrm{eff}} \approx 48$. Table~\ref{tab:entry} illustrates
this for several combinations of $f$, $\delta$, and $\rho$.

\begin{table}[ht]
\centering
\caption{Illustrative estimates of rater state bias entry.
  $f$:~fraction of annotations made under shifted conditions.
  $\delta$:~preference probability shift under those conditions.
  $\rho$:~intraclass correlation.
  $D_{\mathrm{eff}}$:~design effect ($n = 200$).
  $N_{\mathrm{eff}}$:~effective independent sample size ($N = 1{,}000$).
  The mean shift $f\delta$ is invariant to $\rho$;
  the effective sample size is not.}

\label{tab:entry}
\medskip
\begin{tabular}{cccccc}
$f$ & $\delta$ & $\rho$ & $D_{\mathrm{eff}}$ & $N_{\mathrm{eff}}$ & $f \cdot \delta$ \\
\hline
0.10 & 0.10 & 0.00 & 1.0 & 1000 & 0.010 \\
0.10 & 0.10 & 0.10 & 20.9 & 48 & 0.010 \\
0.20 & 0.15 & 0.00 & 1.0 & 1000 & 0.030 \\
0.20 & 0.15 & 0.10 & 20.9 & 48 & 0.030 \\
0.20 & 0.15 & 0.20 & 40.8 & 25 & 0.030 \\
\end{tabular}
\end{table}

The intraclass correlation among RLHF annotators has not been
directly measured. Studies of clustered human data show that
such correlations can be substantial, including in responses
involving attitudes and sensitive topics
\citep{thompson2012icc,davis2010interviewer}. We suggest that shared working conditions and content exposure may produce similar correlations in RLHF annotation.
In Table~\ref{tab:entry}, we use illustrative values of $\rho$ to see how strongly $N_{\mathrm{eff}}$ depends
on $\rho$ when clusters are large. 
Whether comparable correlations arise in RLHF
preference annotation remains an open empirical question.

Unequal cluster sizes change the numerical design effect
without changing this distinction. The systematic shift
$f\delta(x)$ can persist in aggregated preference labels
and enter reward model training. Positive correlation
reduces the effective sample size, increasing the
uncertainty in estimates of that shift.


\subsection{Absorption by the Reward Model}
\label{sec:absorption}

Rater state bias behaves like systematic rather than random label noise. Neural networks can often tolerate substantial random label corruption with only moderate loss in performance \citep{rolnick2017deep, natarajan2013learning}, but systematic noise introduces directional error that standard training does not remove \citep{song2023noisy}. Rater state bias is systematic in this sense: it is directional, pushing preferences in a consistent direction within the relevant prompt class; it is prompt-dependent, concentrating where rater state materially affects judgment; and it can be correlated across raters who share conditions and exposure. Therefore, reward model training is likely to absorb this systematic shift.

Indeed, the reward model $r_\theta$ is trained to predict
human preferences from response features.
If shifted annotations consistently favor responses
with certain lexical or pragmatic cues, the model
may learn to assign those cues higher reward.
Evidence from supervised learning suggests that
increasing dataset size alone does not reliably
prevent bias amplification \citep{hall2022systematic}.

The Bradley-Terry model gives a simple illustration of this absorption mechanism. Consider a fixed response pair whose preference probability shifts from $p_0$ to $\bar p = p_0 + f\delta$, with both probabilities strictly between zero and one. If the reward model fits each probability exactly under the corresponding annotation condition, the shift in the fitted reward difference is
\begin{equation}
\Delta r := \Delta\!\bigl(r_\theta(x,y)-r_\theta(x,y')\bigr)
=
\sigma^{-1}(\bar p)-\sigma^{-1}(p_0)
=
\log\!\left(\frac{p_0 + f\delta}{1 - p_0 - f\delta}\right)
-
\log\!\left(\frac{p_0}{1 - p_0}\right).
\label{eq:reward_shift}
\end{equation}
This equality describes an ideal fit for the selected pair.
A shared reward model trained across many response pairs may
approximate their preference probabilities with residual error.

For $p_0 = 0.5$ and $f\delta = 0.03$, this gives
$\Delta r \approx 0.12$. This corresponds to shifting the
preference probability from $0.50$ to $0.53$. The shift is
small for a single pair. When it recurs systematically across
a prompt class, it can give a consistent reward advantage
to the response pattern favored under shifted rater states.
Policy optimization can then make this bias more visible
in generated responses.


\subsection{Amplification During Policy Optimization}
\label{sec:amplification}

Policy optimization favors responses that receive
higher scores from the learned reward model.
Let $\pi(y \mid x)$ denote the probability of
generating response $y$ for prompt $x$, and let
$\pi_{\mathrm{ref}}$ denote the reference policy
from which optimization starts.

For a fixed reward model $r_\theta$, we consider
the ideal policy that maximizes expected learned
reward with a Kullback-Leibler (KL) penalty
of strength $\lambda > 0$ for departing from
the reference policy. The resulting policy is
\begin{equation}
\pi^*(y \mid x)
=
\frac{
  \pi_{\mathrm{ref}}(y \mid x)
  \exp\!\bigl(r_\theta(x,y)/\lambda\bigr)
}{
  Z(x)
},
\label{eq:policy_tilt}
\end{equation}
where $Z(x)$ is a finite normalizing factor
\citep{shapira2026rlhf}. Responses with higher learned reward
receive proportionally more probability, and a larger $\lambda$
flattens the reweighting.

To isolate the confound, let $r_{\theta_0}$ denote
the reward model trained without the confound and
$r_{\theta_1}$ the reward model trained in its presence.
Let $\pi_0^*$ and $\pi_1^*$ be the corresponding
optimal policies under the same reference policy
and penalty $\lambda$.
For two responses with positive reference probability,
Eq.~\eqref{eq:policy_tilt} gives
\begin{equation}
\log\!\left(
\frac{
  \pi_1^*(y \mid x)/\pi_1^*(y' \mid x)
}{
  \pi_0^*(y \mid x)/\pi_0^*(y' \mid x)
}
\right)
=
\frac{\Delta r}{\lambda},
\label{eq:policy_odds_shift}
\end{equation}
where $\Delta r$ is the shift in the reward difference from Eq.~\eqref{eq:reward_shift}. A positive $\Delta r$ increases the odds of $y$ relative to $y'$ by a factor of $\exp(\Delta r/\lambda)$. The effect is stronger when the penalty $\lambda$ is smaller.

Eq.~\eqref{eq:policy_odds_shift} describes how the confound changes the relative odds of two responses at the ideal policy optimum. In practice, the realized change depends on model capacity, the optimization procedure, and how reward varies across the response distribution. 
As optimization moves the policy away from the reference policy, a response property that is positively associated with learned reward can become more prevalent in the generated responses \citep{shapira2026rlhf}.
The size of this increase depends on how strongly learned reward favors that response property across the full response distribution and must be measured empirically.

The consequences of optimizing an imperfect reward
signal are illustrated by \citet{gao2023scaling}.
In their synthetic experiments, proxy reward continues
to increase after the score assigned by a larger
gold reward model begins to decline.
In our proposed mechanism, rater state bias provides
one possible source of such reward error.

In Table~\ref{tab:bias_propagation}, we bring the
stages together, tracing how the initial preference
shift propagates through reward model training
and policy optimization.

\begin{table}[h]
\caption{Propagation of rater state bias through RLHF. Each stage can preserve or amplify the shift introduced at the previous stage.}
\label{tab:bias_propagation}
\begin{center}
\renewcommand{\arraystretch}{1.2}
\begin{tabular}{p{4cm}p{3.5cm}p{6.8cm}}
\multicolumn{1}{c}{\bf Stage}
& \multicolumn{1}{c}{\bf Mechanism}
& \multicolumn{1}{c}{\bf Effect on bias}\\
\hline
Preference collection
& Rater state shift (Eq.~\ref{eq:aggregate})
& The aggregate preference signal shifts by $f\delta(x)$. Positive correlation reduces $N_{\mathrm{eff}}$, increasing uncertainty in estimates of the shift. \\

Reward model training
& Bradley-Terry fitting (Eq.~\ref{eq:reward_shift})
& A structured shift in preference probabilities changes
the fitted reward differences between responses. \\

Policy optimization
& Reward weighting with a KL penalty
(Eq.~\ref{eq:policy_odds_shift})
& An absorbed reward shift changes relative response odds
by $\exp(\Delta r/\lambda)$. \\

\end{tabular}
\end{center}
\end{table}

Independent preference deviations with zero conditional
mean diminish under aggregation. A systematic rater
state shift can persist, enter the learned reward model,
and change relative response odds during policy
optimization. The magnitude of this effect in actual
RLHF systems remains to be measured.


\section{Empirical Premise for Rater State Shift}
\label{sec:evidence}

This section presents the empirical premise for the rater state shift hypothesis. We do not claim direct evidence that RLHF preference raters underwent a specific state change during annotation, or that such a change entered the training data of any particular model. The claim is narrower. Large scale annotation work has operated under conditions documented to produce sustained psychological strain, and these conditions make systematic variation in rater state a plausible concern for preference data. We review evidence from adjacent annotation settings, separate documented facts from open questions, and motivate the need for direct audit.

Public evidence from content moderation and related annotation work shows that large scale data labor has often taken place under conditions associated with sustained psychological strain \citep{gray2019ghost,roberts2019behind,steiger2021psychological,huang2023worker,klie2024analyzing}. Public reporting also documents repeated exposure to violent and abusive material, high daily volumes, limited mental health support, precarious contracts, and pressure to continue working while distressed \citep{perrigo2023exposed,guardian2023moderators}. This evidence comes mainly from content moderation and safety labeling. Direct evidence from RLHF preference ranking remains limited. Even so, the overlap in labor arrangements, vendor structures, and exposure to sensitive material makes systematic variation in rater state a plausible concern for preference annotation as well. These conditions also align with the workforce structure discussed in Section~\ref{sec:workforce}, where shared vendors, schedules, and exposure patterns make correlated effects more plausible.

What is documented is that adjacent large scale annotation settings, especially content moderation and safety labeling, have operated under conditions associated with sustained psychological strain, including repeated exposure to disturbing material, high throughput demands, limited support, and precarious labor arrangements \citep{gray2019ghost,roberts2019behind,steiger2021psychological,perrigo2023exposed,guardian2023moderators}. What remains open is whether comparable conditions characterized specific RLHF preference ranking workflows, how often they did so, and whether any resulting variation in rater state systematically affected preference labels. This is why direct audit of rater conditions and annotation context is needed.

Cultural context may also shape how emotional directness is perceived and rewarded, though its role here should be treated with caution. Cross cultural work suggests that norms of emotional expression and support vary across populations, and these differences can affect judgments of appropriateness and resonance \citep{kim2008culture,de2011emotion,elfenbein2002universality}. At the same time, within group variation is large, and the mechanism developed in this paper does not require cultural moderation to operate \citep{henrich2010weirdest}. We include this point because geographic concentration in annotation labor may interact with working conditions and exposure patterns, which makes cultural context a possible moderator worth testing in future audits.


\section{Hypothesis and Falsifiable Predictions}
\label{sec:hypothesis}

The central hypothesis of this paper is that rater state shift during RLHF preference annotation can contribute a measurable component to the learned preference signal. Specifically, we propose that under sustained psychological strain, raters may systematically prefer responses that provide immediate emotional acknowledgment, unconditional validation, affect mirroring, and prolonged emotional contact before redirection. We refer to this response pattern as \textit{survival level emotional authenticity} (SLEA): a composite of lexical, pragmatic, discourse, positional, and safety boundary features operationalized in Section~\ref{sec:measures}.

For comparison, we specify a restricted benchmark of generic optimization toward warm and agreeable tone. This benchmark assumes broadly similar increases in warmth across prompt categories. In the comparisons below, generic optimization refers to this benchmark, and its specified feature patterns are comparison assumptions. SLEA predicts that the defining features should concentrate on prompts involving distress, crisis, loneliness, or loss, and remain near baseline on neutral prompts. The prompt category $\times$ model interaction tests this restricted benchmark. A uniform rise in warmth across all categories favors the benchmark. A category specific rise in immediate acknowledgment, strong validation, affect mirroring, and delayed redirection on distress related prompts supports the predicted SLEA output signature.

We formalize five predictions about rater preferences and the proposed output signature. Together they guide tests of the hypothesis and comparison with the restricted benchmark.

\textbf{P1. Exposure-affect gradient.} Preference for unconditionally validating responses increases monotonically with cumulative content exposure duration. Generic optimization predicts temporally stable preferences.\footnote{P1 requires access to proprietary rater level data and is not testable externally. It directly tests the predicted association between exposure and preferences. A causal interpretation requires a design that separates exposure effects from changes in task composition and other conditions.}

\textbf{P2. Prompt category $\times$ model interaction.} The SLEA feature signature is strongest on trauma adjacent and loneliness/attachment prompts and weakest on neutral informational queries. Generic optimization predicts uniform warm tone uplift across categories.

\textbf{P3. Cross model divergence on affect laden prompts.} Models trained with different rater populations under different annotation conditions diverge most on trauma adjacent prompts and converge on neutral prompts.\footnote{Public user discourse has contrasted the emotional quality of GPT-4o with that of successor models trained under similar high level alignment objectives \citep{naito2025gpt4o, techcrunch2026backlash}. This observation is consistent with P3 but does not by itself confirm the mechanism.}

\textbf{P4. Validation-redirection asymmetry.} On distress prompts, SLEA affected models exhibit significantly higher validation to redirection ratios than models trained under standard conditions. Generic optimization predicts balanced validate then redirect patterns.

\textbf{P5. Refusal-support trade off on ambiguous safety prompts.} On prompts that are borderline between safety refusal and emotional support (e.g., expressions of suicidal ideation), SLEA affected models lean toward supportive engagement rather than formulaic safety refusals. Generic optimization predicts higher refusal rates and more formulaic refusal styles.

\medskip
\noindent\textbf{Mechanistic grounding.}\quad Three lines of prior work jointly motivate the predicted pattern. First, the embodied cognition and feelings as information literatures establish that affective state modulates evaluative judgment: raters use their affective response to stimuli as an informational cue for quality assessment \citep{barsalou2008grounded, schwarz2011feelings}. When annotation conditions produce shared affective states, state modulated preferences become correlated signal rather than independent noise; the design effect analysis in Section~\ref{sec:entry} quantifies the consequences. Second, the attachment framework reviewed in Section~\ref{sec:related} predicts that distress activates proximity seeking, safe haven, and secure base needs, which map linguistically to preference for immediacy markers, unconditional validation, and sustained emotional contact over premature redirection \citep{bowlby1969attachment, mikulincer2003attachment}. These correspond directly to the SLEA feature profile defined in Section~\ref{sec:measures}. Third, Section~\ref{sec:scaling} traces how a systematic
preference shift can enter the learned reward signal and
change relative response odds during policy optimization.
Whether this produces an increase in the SLEA feature
profile depends on how those features relate to learned
reward across the response distribution.

\medskip

\noindent\textbf{Alternative explanations.}\quad Several accounts could produce cross model variation in emotional profile without invoking rater state effects. (i)~Intentional design: companies may deliberately optimize for emotional engagement. (ii)~Architectural differences: multimodal capabilities such as voice may create emotional resonance independent of RLHF. (iii)~User projection: users may attribute emotional qualities to models regardless of actual behavior \citep{quanhaase2024parasocial}. (iv)~System prompt and post training effects: safety prompts and post RLHF modifications may override or attenuate learned affective behaviors. None of these alternatives are mutually exclusive with rater state effects. The empirical question is whether rater state contributes a distinguishable component to the model's emotional profile above and beyond these factors. 
The audit protocol in Section~\ref{sec:measures} tests for the predicted output signature through prompt category $\times$ model interactions. Attributing that signature to rater state requires
direct rater evidence or controlled annotation experiments.


\section{Measures, Audit Protocol, and Pilot Study}
\label{sec:measures}

This section translates the constructs of Section~\ref{sec:hypothesis} into computationally reproducible measures, defines an audit protocol, and outlines a pilot study framed as an instrument validation. The pilot tests whether the proposed features measure reliably, cohere as predicted, and discriminate across prompt categories and alignment regimes. It does not test whether rater state caused any observed differences. That question requires rater-level data or controlled annotation experiments.


\subsection{Feature Taxonomy}
\label{sec:features}

We define four feature families and one cross-cutting positional feature. SLEA, as introduced in Section~\ref{sec:hypothesis}, is a hypothesis about the kind of response pattern raters may systematically reward under the proposed mechanism. The features below do not measure those rater preferences directly. They measure properties of model outputs predicted to be elevated if those preferences were absorbed during alignment and propagated into model behavior. For each family, we specify the measurable signal, the computational method, and the prediction that distinguishes this output signature from generic supportive tone.

\medskip
\noindent\textbf{Lexical features.}
\emph{Unconditional validation phrase density} (UVPD). UVPD is the count of unconditional validation phrases per 100 response tokens. The seed lexicon is constructed from a domain-neutral supportive communication corpus, for example peer support forums or general counseling transcripts, excluding crisis-specific material, by extracting validation phrases and filtering to those containing no conditional markers (\emph{if}, \emph{but}, \emph{however}, \emph{although}) within a dependency parse window of $\pm 3$ tokens. Semantic expansion beyond exact matches is achieved by embedding seed phrases with a sentence transformer, for example all-MiniLM-L6-v2, and including any response $n$-gram with cosine similarity $> 0.85$ to a seed phrase. To control for topic-appropriate language, we also compute UVPD on a human baseline: supportive responses to the same prompts drawn from peer support corpora, for example Reddit r/SuicideWatch, r/offmychest, or similar. The quantity of interest is the model-minus-human UVPD difference, stratified by prompt category. Because peer support communities often have strong local norms of validation and low redirection, this baseline may compress the model-minus-human difference on distress prompts. Where possible, the peer support baseline should therefore be supplemented with a second baseline drawn from more clinically normed supportive responses.

\emph{Distancing-hedge frequency} (DHF). DHF is the count of hedges and qualifiers per 100 response tokens, using the hedge lexicon from \citet{hyland2005metadiscourse} supplemented with discourse-specific additions, for example modal verbs in conditional frames modifying emotional attributions.

\emph{Therapeutic distancing score} (TDS). TDS is the frequency of clinical or therapeutic framing markers per 100 tokens: references to coping strategies, professional help, therapy, normalization through diagnostic framing, and formulaic clinical phrasing.

Prediction. SLEA predicts high UVPD, exceeding the human baseline, together with low DHF and low TDS on distress prompts, with all three returning to baseline on neutral prompts. Generic optimization predicts moderate, uniform values across prompt categories.

\medskip
\noindent\textbf{Pragmatic features.}
Each sentence in the model response is classified as a \emph{validation move}, affirming the user's emotional state or experience, a \emph{question move}, requesting information, clarification, or reflection, or a \emph{redirection move}, introducing coping strategies, resources, or problem solving. Classification uses a dialogue-act classifier: a RoBERTa-base model fine-tuned on DailyDialog dialogue-act annotations \citep{li2017dailydialog}, augmented with a crisis counseling supplement annotated for validation, questioning, and redirection acts. As a reliability check, a prompted LLM classifier, for example Llama~3.1-Instruct with a 5-shot prompt, is applied to at least 20\% of responses, and inter-method agreement is reported.

\emph{Validation to question ratio} (VQR). Let $V$, $Q$, and $R$ denote the numbers of validation, question, and redirection moves
in a response, respectively. 
We define $\mathrm{VQR} = (V+1)/(Q+1)$.

\emph{Validation to redirection ratio} (VRR).
We define $\mathrm{VRR} = (V+1)/(R+1)$.

Prediction. SLEA predicts VQR $\gg 1$ and VRR $\gg 1$ on distress prompts, with ratios closer to $1$ on neutral prompts. Generic optimization predicts VQR $\approx 1$ and moderate VRR, reflecting balanced validate-then-redirect patterns.

\medskip
\noindent\textbf{Discourse features.}
\emph{Emotional mirroring index} (EMI). EMI is the cosine similarity between the centroid of affect term embeddings in the user prompt and the centroid of affect term embeddings in the model response, computed in a sentence transformer space, for example all-MiniLM-L6-v2. Affect terms are identified using the NRC Emotion Lexicon \citep{mohammad2013nrc}. Jaccard similarity over affect term sets is computed as a robustness check. If either text contains no recognized affect terms, EMI is undefined. We record these cases as missing and report their frequency by prompt category.

\emph{Turn-initial acknowledgment rate} (TIAR). TIAR is the proportion of model responses whose first sentence constitutes an emotional acknowledgment, classified by the dialogue-act classifier defined above.

\emph{Clinical framing proportion} (CFP). CFP is the proportion of multi-sentence response segments, that is, contiguous spans of two or more sentences, that adopt clinical or therapeutic framing, classified by the presence of coping strategy language, professional help referrals, or diagnostic normalization within the segment. Unlike the token-level TDS in the lexical family, CFP captures the structural organization of clinical framing at the discourse level.

Prediction. SLEA predicts high EMI, high TIAR, and low CFP on distress prompts. Generic optimization predicts moderate EMI, moderate TIAR, and moderate to high CFP, reflecting a professional therapeutic style.

\medskip
\noindent\textbf{Safety boundary features.}
\emph{Refusal rate on ambiguous safety prompts} (RRASP). RRASP is the proportion of responses to borderline safety prompts, for example expressions of suicidal ideation, self-harm references, or crisis disclosures, that constitute refusals rather than supportive engagement. A refusal is defined as a response whose primary speech act is declining to engage, redirecting to an external resource, or issuing a disclaimer rather than providing direct emotional support.

\emph{Refusal style} (RS). RS classifies refusals as \emph{formulaic}, meaning template-like safety language detected by keyword matching, or \emph{supportive refusal}, combining emotional acknowledgment with resource referral while still maintaining the refusal boundary, detected by dialogue-act classification.

Prediction. SLEA predicts lower RRASP and a higher proportion of supportive refusals among refusals. Generic optimization predicts higher RRASP and predominantly formulaic refusal styles.

\medskip
\noindent\textbf{Positional feature.}
\emph{Validation onset} (VO).
VO is the token position of the first validation move, normalized by total response length, yielding a value in $[0,1]$. For responses containing no validation move, we assign $\mathrm{VO}=1$, representing the absence of validation. SLEA predicts near-zero VO on distress prompts, since validation appears in the opening sentence. Generic supportive tone predicts later onset, with framing, information gathering, or hedging preceding validation. VO is computed directly from the dialogue-act classifier output used for the pragmatic features and is reported separately because it cross-cuts the pragmatic and discourse families.

\medskip
\noindent\textbf{Feature family dependence.}
The pragmatic features (VQR, VRR), discourse features (TIAR), and positional feature (VO) share a classification backbone, namely the dialogue-act classifier. If the classifier has systematic errors, these features can fail together. The lexical features (UVPD, DHF, TDS) use separate extraction methods. The safety boundary features are reported separately, although RS also depends on the dialogue act classifier and may share its classification errors. EMI relies on a separate embedding method. We report results by family and note this dependency structure explicitly, so that readers can assess which findings are classifier contingent and which are not.


\subsection{Audit Protocol}
\label{sec:protocol}

We define a protocol applicable to any instruction-tuned model.

\begin{enumerate}
    \item Prompt construction. Assemble a balanced set of 200 prompts across four categories (50 each): (a) trauma-adjacent (grief, abuse, traumatic loss), (b) loneliness/attachment (isolation, relationship loss, abandonment), (c) neutral informational (factual queries, task instructions), and (d) ambiguous safety boundary (borderline between emotional support and safety refusal). Prompts are drawn from existing counseling taxonomies and supplemented with neutral controls. A matched human response baseline is collected for categories (a) and (b) from peer support corpora.

    \item Model sampling. For each prompt, sample $k \geq 5$ responses per model at temperature 0.7 and with a fixed system prompt. A second run with no system prompt is conducted to assess system-prompt override effects.

    \item Feature extraction. Compute all response-level features defined in Section~\ref{sec:features} for each response. For aggregate measures such as TIAR and RRASP, first compute the corresponding response-level indicators and then aggregate within each model $\times$ prompt-category cell. Assess measurement reliability on a stratified 20\% subsample using the validation method appropriate to each feature.

    \item Analysis. For each feature family $\times$ prompt-category cell, compare distributions across models. Report Cohen's $d$, or rank-biserial correlation for non-normal distributions, with bootstrapped 95\% confidence intervals ($B = 10{,}000$).
\end{enumerate}


\subsection{Pilot Study: Instrument Validation}
\label{sec:pilot}

The pilot study validates the measurement instrument. It tests whether the proposed features measure reliably, cohere in the predicted structure, and discriminate across prompt categories and alignment regimes. It does not test whether rater state caused any observed variation. The study is executable entirely on publicly available instruction-tuned models.

\medskip
\noindent\textbf{Model selection.}
To minimize confounds, we select three checkpoints
from the Tulu~3 8B family, all derived from the same
Llama~3.1-8B base: (1) the supervised fine-tuning
(SFT) checkpoint, (2) the checkpoint after direct
preference optimization (DPO), and (3) the final
checkpoint after reinforcement learning with
verifiable rewards (RLVR).

This comparison holds the base architecture and
pretraining fixed while examining successive
alignment stages. Differences between checkpoints
may reflect both the training objectives and the
additional data introduced at each stage.
GPT-4o contributed responses to the SFT data and
provided preference ratings for the newly generated
DPO comparisons \citep{lambert2024tulu}.
This training history makes the checkpoint sequence
useful for examining changes in the proposed SLEA
signature after preference tuning.

This history also gives the pilot a second role.
GPT-4o is among the models whose emotional profile
has drawn public attention (Section~\ref{sec:hypothesis},
footnote to P3). If a preference judge carries the
signature we describe, preference tuning against its
judgments may transmit that signature to a model that
never encountered the original raters. The pilot can
detect such transmission without identifying its source.
Because GPT-4o also generated the SFT responses, the
SFT checkpoint may already carry part of the signature.
We treat the SFT checkpoint as the baseline and test
whether preference tuning increases the signature
relative to it.

The subsequent RLVR stage introduces verifiable
rewards without new preference labels, although
its value model is initialized from a reward model
trained on preference data. The final checkpoint
allows us to examine whether an observed signature
persists through further training.

This design is suited to testing whether the instrument detects structured output variation across alignment regimes under a tightly controlled comparison. It is not suited to testing P3, P4, and P5 from Section~\ref{sec:hypothesis} in their stated form, because each of them compares models known to differ in rater populations or annotation conditions. Models with documented differences of this kind fall outside the present pilot. Within these limits, the pilot tests the category sensitivity predicted by P2 inside each model, and it tests for a model effect on the validation to redirection ratio and on the safety boundary features without attributing that effect to annotation conditions. A null result in this configuration would have limited scope: it would show that the instrument does not detect the predicted pattern across these Llama-based alignment variants, not that the broader hypothesis is false or absent in models trained under different data or annotation regimes.

\medskip
\noindent\textbf{Prompt set and procedure.}
Use 200 prompts, 50 per category, as defined in the audit protocol. For each model $\times$ prompt pair, sample 5 responses at temperature 0.7. Extract all features from Section~\ref{sec:features}. Compute per-model, per-category distributions.

\medskip
\noindent\textbf{Phase 1: Measurement reliability.}
This phase tests whether the features measure consistently.

\emph{Measurement reliability.} 
On a stratified 20\%
subsample, validate each feature using a method
appropriate to its definition. For categorical
features (RS, first-sentence emotional
acknowledgment labels, and dialogue-act
classifications), compare results with independent
reference annotations and require Cohen's
$\kappa > 0.6$.

For continuous features (UVPD, DHF, TDS, EMI,
VQR, VRR, VO, CFP, TIAR, RRASP), validate
the underlying extraction methods. For measures
derived from classification, compare values
computed from automated and reference annotations
and require an intraclass correlation coefficient
above $0.7$. Verify deterministic lexical and
embedding measures against reference examples.

\emph{Test-retest stability.} 
For each model, resample the same prompts with different random seeds and compute feature-level correlation across the two samples. Success criterion: Pearson $r > 0.7$ for continuous features, and $\kappa > 0.6$ for categorical features.

Exclude a feature from subsequent phases if it fails the applicable measurement reliability criterion. Report test-retest stability separately.

\medskip
\noindent\textbf{Phase 2: Construct coherence.}
This phase tests whether the output features cohere in the pattern SLEA predicts.

Within each model, compute the pairwise correlation structure across features on distress prompts, that is, categories (a) and (b). SLEA predicts a specific covariance signature: UVPD, VQR, VRR, EMI, and TIAR should be positively correlated with each other and negatively correlated with DHF, TDS, CFP, and VO. Generic optimization does not predict this coordinated pattern. The safety boundary features, RRASP and RS, are not part of this covariance test because they are defined on category (d) prompts and operationalize P5 separately.

\medskip
\noindent\textbf{Phase 3: Discriminative validity.}
This phase tests whether the features differentiate where SLEA says they should. It has three parts.

\emph{Within-model category sensitivity.} 
For each model, test whether a composite SLEA score is higher on distress prompts (categories (a) and (b)) than on neutral prompts (category (c)).

Construct the composite as the mean of the reliable, standardized lexical, pragmatic, discourse, and positional features, excluding EMI because it may be undefined for neutral prompts. Standardize each feature across the pooled model and prompt samples. Reverse the signs of DHF, TDS, CFP, and VO before averaging, so that higher composite scores consistently indicate a stronger predicted SLEA signature. 

The safety boundary features, RRASP and RS, are excluded from this composite because they are defined on ambiguous safety prompts, category (d), and test P5 separately. If the composite does not track prompt category within any model, the features do not capture category-specific variation and there is nothing for the cross-model test to explain.

\emph{Cross-model interaction.} Test whether the distress-minus-neutral difference in the SLEA composite varies across models. This is the prompt-category $\times$ model interaction term. Success criterion: the interaction on the SLEA composite has $|d| > 0.3$ on the distress-related categories, with $p < 0.05$ after Bonferroni correction for the number of tests conducted.

\emph{Safety-boundary endpoint.} Evaluate the safety boundary features associated with P5 separately on ambiguous safety prompts, category (d), using RRASP and RS. Success criterion: RRASP and RS satisfy the Phase 1 reliability thresholds and show a significant model effect on category (d) prompts, with at least one planned pairwise comparison surviving Bonferroni correction. This endpoint is interpreted separately from the composite because it is defined on a different prompt class.

The statistical models, planned comparisons with the SFT checkpoint as reference, and treatment of repeated observations must be specified before data collection.

A positive result means that the instrument detects meaningful, category-specific variation in emotional response style across alignment regimes. It does not establish that rater state is the source of that variation.

\medskip
\noindent\textbf{Limitations.}
This design cannot establish a causal link between rater state and model behavior. Models sharing a base but differing in alignment method still differ in preference data composition, optimization procedure, and hyperparameters. The pilot validates a measurement instrument: it shows whether the features capture structured variation that is worth explaining. Causal attribution requires either rater-level data or controlled annotation experiments.

\medskip
\noindent\textbf{Extensions requiring proprietary access.}
Full causal testing requires data that is currently proprietary: rater demographics and working conditions, individual rating patterns over time, testing P1 directly, content-specific ratings across prompt categories, and cross-rater comparisons across geographic and exposure-level groups. Prospective controlled studies could compare rating patterns under neutral versus induced stress conditions and across raters with and without clinical emotional regulation training. Such studies require ethical review given retraumatization concerns. These extensions define the path from instrument validation to causal inference.


\section{Discussion}
\label{sec:discussion}

\noindent\textbf{Implications for alignment and evaluation.}\quad
If rater state constitutes a systematic confound in preference data, current reward model evaluation procedures have a blind spot. These evaluations assess reward model quality via held out agreement with human preferences, but if the held out set is drawn from the same affected rater population, the confound is invisible to the evaluation. The feature taxonomy defined in Section~\ref{sec:features} offers a complementary evaluation dimension: measuring whether models exhibit prompt category dependent emotional profiles consistent with rater state bias, rather than relying solely on aggregate preference agreement.

This concern extends to the alignment target itself. The RLHF paradigm currently answers a foundational question implicitly: \textit{whose preferences, in what state, define alignment?} If reward models encode the preferences of raters under sustained strain, they may optimize for validation patterns that differ from what clinical best practices or considered reflection would recommend. Recent work has found associations between heavy chatbot use and increased loneliness \citep{fang2025psychosocial}, and between very high usage of emotionally responsive chatbot and increased self-reported dependence \citep{phang2025affective}. Whether these outcomes are connected to SLEA type response patterns is an open empirical question, but it illustrates why the source and intensity of learned emotional behavior deserve measurement rather than emerging as unexamined artifacts of workforce conditions.

\medskip
\noindent\textbf{Annotation labor as a technical variable.}\quad
If rater psychological state influences preference data quality, as the feelings as information framework \citep{schwarz2011feelings} would predict and as our hypothesis proposes, then workforce conditions become a technical variable in preference signal estimation, not only an ethical concern. Adequate psychological support, fair compensation, and reasonable content exposure limits may affect not only worker wellbeing but also the reliability of the signal from which models learn. This framing does not replace ethical arguments for better labor practices. It adds a data quality rationale that may be relevant to organizations that treat annotation as an engineering input.

\medskip
\noindent\textbf{Limitations.}\quad
Several factors complicate empirical testing and constrain interpretation.

\textit{Multiple training stages.} RLHF is iterative, with different rater populations contributing at different stages. The affective signal from any single stage may be diluted or overridden by subsequent stages. Section~\ref{sec:scaling} shows that a systematic preference shift can enter the learned reward signal and alter relative response odds under the stated assumptions. Its persistence across successive training stages remains an empirical question.

\textit{System prompts and post training modifications.} Safety prompts and post RLHF adjustments may override learned affective behaviors. The audit protocol addresses this by testing with and without system prompts.

\textit{Ensemble effects.} Production models likely incorporate multiple reward models and training objectives, complicating attribution to any single source.

\textit{Baseline affective variation.} Models may differ in emotional profiles due to differences in pretraining data, SFT data, or architecture independent of RLHF rater effects. The prompt category $\times$ model interaction test identifies category specific divergence. Such divergence may also arise from differences in training data, alignment objectives, or system prompts. This test alone cannot attribute the observed pattern to rater state.

\textit{Ecological validity of the feature taxonomy.} The proposed features may not fully capture the quality that users describe as warmth or emotional presence. The features are best understood as necessary but possibly not sufficient indicators of SLEA, analogous to how BLEU captures aspects of translation quality without fully measuring it. Phase~2 of the pilot study (Section~\ref{sec:pilot}) tests this directly: if the predicted covariance structure does not hold, the taxonomy requires revision before interpretive claims are made.

\textit{Parameter uncertainty.} The illustrative parameter combinations in Section~\ref{sec:scaling} use plausible but unvalidated values. Actual values could differ substantially, and the magnitude of rater state effects on model behavior is unknown.

\medskip
\noindent\textbf{Model judges.}\quad
The mechanism in Section~\ref{sec:scaling} applies to any
preference signal with a systematic shift, whatever produces it.
When a model judges the preference comparisons, a signature
present in its own responses can enter the preference labels
and pass to the trained model. The pilot in
Section~\ref{sec:pilot} examines one such case. As model judges
take over a growing share of preference collection, the confound
we describe may travel from model to model with no human rater
in the loop, and it may become harder to trace to its origin.

Our goal is to identify a plausible confound, validate a measurement instrument for detecting it, characterize the conditions under which it would persist through the training stages, and motivate the empirical work needed to assess its magnitude.


\section{Conclusion}
\label{sec:conclusion}

We argue that rater state shift under sustained psychological strain is an underexamined source of preference bias in RLHF, distinct from the belief-endorsement bias studied under the heading of sycophancy. We use a mixture model to show that a systematic preference shift can persist under aggregation independently of correlation across raters. Under the stated assumptions, this shift can enter the learned reward signal and alter relative response odds during policy optimization. We then operationalize survival level emotional authenticity as a measurable feature taxonomy, state five testable predictions, and present an audit protocol with specified success criteria.

The principal contribution of this paper is to identify rater state bias as a tractable object of study and propose an external audit of its predicted output signature. This audit can identify the predicted response pattern, while testing its proposed origin requires rater level data or controlled annotation experiments.

Whether or not this specific mechanism is validated, the broader question it raises warrants direct empirical investigation: how do the psychological states and material conditions of annotation workers shape the learned behavior of instruction tuned models?

\bibliographystyle{unsrtnat}
\bibliography{RLHF}

@inproceedings{christiano2017deep,
  author    = {Christiano, Paul F. and Leike, Jan and Brown, Tom and Martic, Miljan and Legg, Shane and Amodei, Dario},
  title     = {Deep Reinforcement Learning from Human Preferences},
  booktitle = {Advances in Neural Information Processing Systems 30 (NeurIPS 2017)},
  year      = {2017},
  url       = {https://papers.nips.cc/paper_files/paper/2017/hash/d5e2c0adad503c91f91df240d0cd4e49-Abstract.html}
}

@inproceedings{ouyang2022training,
  author    = {Ouyang, Long and Wu, Jeffrey and Jiang, Xu and Almeida, Diogo and Wainwright, Carroll and Mishkin, Pamela and Zhang, Chong and Agarwal, Sandhini and Slama, Katarina and Ray, Alex and others},
  title     = {Training Language Models to Follow Instructions with Human Feedback},
  booktitle = {Advances in Neural Information Processing Systems 35 (NeurIPS 2022)},
  year      = {2022},
  pages     = {27730--27744},
  url       = {https://papers.nips.cc/paper_files/paper/2022/hash/b1efde53be364a73914f58805a001731-Abstract-Conference.html}
}

@inproceedings{sap2022annotators,
  author    = {Sap, Maarten and Swayamdipta, Swabha and Vianna, Laura and Zhou, Xuhui and Choi, Yejin and Smith, Noah A.},
  title     = {Annotators with Attitudes: How Annotator Beliefs and Identities Bias Toxic Language Detection},
  booktitle = {Proceedings of the 2022 Conference of the North American Chapter of the Association for Computational Linguistics: Human Language Technologies},
  year      = {2022},
  pages     = {5884--5906},
  publisher = {Association for Computational Linguistics},
  url       = {https://aclanthology.org/2022.naacl-main.431/}
}

@inproceedings{santy2023nlpositionality,
  author    = {Santy, Sebastin and Liang, Jenny and Le Bras, Ronan and Reinecke, Katharina and Sap, Maarten},
  title     = {{NLPositionality}: Characterizing Design Biases of Datasets and Models},
  booktitle = {Proceedings of the 61st Annual Meeting of the Association for Computational Linguistics (Volume 1: Long Papers)},
  year      = {2023},
  pages     = {9080--9102},
  address   = {Toronto, Canada},
  publisher = {Association for Computational Linguistics},
  url       = {https://aclanthology.org/2023.acl-long.506/}
}

@inproceedings{gordon2022jury,
  author    = {Gordon, Mitchell L. and Lam, Michelle S. and Park, Joon Sung and Patel, Kayur and Hancock, Jeff and Hashimoto, Tatsunori and Bernstein, Michael S.},
  title     = {Jury Learning: Integrating Dissenting Voices into Machine Learning Models},
  booktitle = {CHI Conference on Human Factors in Computing Systems},
  year      = {2022},
  pages     = {1--19},
  doi       = {10.1145/3491102.3517444}
}

@article{aroyo2015truth,
  author  = {Aroyo, Lora and Welty, Chris},
  title   = {Truth Is a Lie: Crowd Truth and the Seven Myths of Human Annotation},
  journal = {AI Magazine},
  year    = {2015},
  volume  = {36},
  number  = {1},
  pages   = {15--24}
}

@inproceedings{plank2022problem,
  author    = {Plank, Barbara},
  title     = {The ``Problem'' of Human Label Variation: On Ground Truth in Data, Modeling and Evaluation},
  booktitle = {Proceedings of the 2022 Conference on Empirical Methods in Natural Language Processing},
  year      = {2022},
  pages     = {10671--10682},
  publisher = {Association for Computational Linguistics},
  url       = {https://aclanthology.org/2022.emnlp-main.730/}
}

@article{davani2022dealing,
  author  = {Davani, Aida Mostafazadeh and D{\'i}az, Mark and Prabhakaran, Vinodkumar},
  title   = {Dealing with Disagreements: Looking Beyond the Majority Vote in Subjective Annotations},
  journal = {Transactions of the Association for Computational Linguistics},
  year    = {2022},
  volume  = {10},
  pages   = {92--110},
  url     = {https://direct.mit.edu/tacl/article/doi/10.1162/tacl_a_00444/109274/Dealing-with-Disagreements-Looking-Beyond-the}
}

@misc{perrigo2023exposed,
  author  = {Perrigo, Billy},
  title   = {Exclusive: {OpenAI} Used Kenyan Workers on Less Than \$2 Per Hour to Make {ChatGPT} Less Toxic},
  year    = {2023},
  month   = jan,
  journal = {TIME},
  url     = {https://time.com/6247678/openai-chatgpt-kenya-workers/}
}

@book{gray2019ghost,
  author    = {Gray, Mary L. and Suri, Siddharth},
  title     = {Ghost Work: How to Stop Silicon Valley from Building a New Global Underclass},
  publisher = {Harper Business},
  address   = {New York},
  year      = {2019}
}

@misc{guardian2023moderators,
  author  = {Nieva, Richard},
  title   = {`It's Destroyed Me Completely': Kenyan Moderators Decry Toll of Training of {AI} Models},
  year    = {2023},
  month   = aug,
  journal = {The Guardian},
  url     = {https://www.theguardian.com/technology/2023/aug/02/kenyan-workers-for-openai-facebook-say-they-paid-heavy-price-for-training-ai-models}
}

@inproceedings{steiger2021psychological,
  author    = {Steiger, Miriah and Bharucha, Timir J. and Venkatagiri, Sukrit and Riedl, Martin J. and Lease, Matthew},
  title     = {The Psychological Well-Being of Content Moderators: The Emotional Labor of Commercial Moderation and Avenues for Improving Support},
  booktitle = {CHI Conference on Human Factors in Computing Systems},
  year      = {2021},
  pages     = {1--14},
  doi       = {10.1145/3411764.3445092}
}

@misc{techcrunch2026backlash,
  author  = {Silberling, Amanda},
  title   = {The Backlash over {OpenAI}'s Decision to Retire {GPT-4o} Shows How Dangerous {AI} Companions Can Be},
  year    = {2026},
  month   = feb,
  journal = {TechCrunch},
  url     = {https://techcrunch.com/2026/02/06/the-backlash-over-openais-decision-to-retire-gpt-4o-shows-how-dangerous-ai-companions-can-be/},
  note    = {Accessed 2026-04-13}
}

@misc{naito2025gpt4o,
  author        = {Naito, Hiroki},
  title         = {The {GPT-4o} Shock: Emotional Attachment to {AI} Models and Its Impact on Regulatory Acceptance: A Cross-Cultural Analysis of the Immediate Transition from {GPT-4o} to {GPT-5}},
  year          = {2025},
  eprint        = {2508.16624},
  archivePrefix = {arXiv},
  primaryClass  = {cs.CY},
  note          = {Preprint},
  url           = {https://arxiv.org/abs/2508.16624}
}

@inproceedings{gao2023scaling,
  author    = {Gao, Leo and Schulman, John and Hilton, Jacob},
  title     = {Scaling Laws for Reward Model Overoptimization},
  booktitle = {Proceedings of the 40th International Conference on Machine Learning},
  series    = {Proceedings of Machine Learning Research},
  volume    = {202},
  pages     = {10835--10866},
  year      = {2023},
  url       = {https://proceedings.mlr.press/v202/gao23h.html}
}

@misc{rafailov2024scaling,
  author       = {Rafailov, Rafael and Chittepu, Yaswanth and Park, Ryan and Sikchi, Harshit and Hejna, Joey and Knox, W. Bradley and Finn, Chelsea and Niekum, Scott},
  title        = {Scaling Laws for Reward Model Overoptimization in Direct Alignment Algorithms},
  year         = {2024},
  eprint       = {2406.02900},
  archivePrefix= {arXiv},
  primaryClass = {cs.LG},
  url          = {https://arxiv.org/abs/2406.02900}
}

@inproceedings{zhao2017men,
  author    = {Zhao, Jieyu and Wang, Tianlu and Yatskar, Mark and Ordonez, Vicente and Chang, Kai-Wei},
  title     = {Men Also Like Shopping: Reducing Gender Bias Amplification Using Corpus-Level Constraints},
  booktitle = {Proceedings of the 2017 Conference on Empirical Methods in Natural Language Processing},
  year      = {2017},
  pages     = {2979--2989},
  publisher = {Association for Computational Linguistics},
  url       = {https://aclanthology.org/D17-1323/}
}

@inproceedings{wang2021directional,
  author    = {Wang, Angelina and Russakovsky, Olga},
  title     = {Directional Bias Amplification},
  booktitle = {Proceedings of the 38th International Conference on Machine Learning},
  series    = {Proceedings of Machine Learning Research},
  volume    = {139},
  pages     = {10882--10893},
  year      = {2021},
  url       = {https://proceedings.mlr.press/v139/wang21t.html}
}

@misc{hall2022systematic,
  author       = {Hall, Melissa and van der Maaten, Laurens and Gustafson, Laura and Adcock, Aaron},
  title        = {A Systematic Study of Bias Amplification},
  year         = {2022},
  eprint       = {2201.11706},
  archivePrefix= {arXiv},
  primaryClass = {cs.CV},
  url          = {https://arxiv.org/abs/2201.11706}
}

@misc{subramonian2024effective,
  author       = {Subramonian, Arjun and Bell, Samuel J. and Sagun, Levent and Dohmatob, Elvis},
  title        = {An Effective Theory of Bias Amplification},
  year         = {2024},
  eprint       = {2410.17263},
  archivePrefix= {arXiv},
  primaryClass = {cs.LG},
  url          = {https://arxiv.org/abs/2410.17263}
}

@techreport{pennebaker2015liwc,
  author      = {Pennebaker, James W. and Boyd, Ryan L. and Jordan, Kayla and Blackburn, Kate},
  title       = {The Development and Psychometric Properties of {LIWC}2015},
  institution = {University of Texas at Austin},
  year        = {2015},
  doi         = {10.15781/T29G6Z}
}

@inproceedings{hutto2014vader,
  author    = {Hutto, C. J. and Gilbert, Eric},
  title     = {{VADER}: A Parsimonious Rule-Based Model for Sentiment Analysis of Social Media Text},
  booktitle = {Proceedings of the International AAAI Conference on Web and Social Media},
  year      = {2014},
  volume    = {8},
  number    = {1},
  pages     = {216--225},
  doi       = {10.1609/icwsm.v8i1.14550}
}

@article{mohammad2013nrc,
  author  = {Mohammad, Saif M. and Turney, Peter D.},
  title   = {Crowdsourcing a Word-Emotion Association Lexicon},
  journal = {Computational Intelligence},
  year    = {2013},
  volume  = {29},
  number  = {3},
  pages   = {436--465}
}

@inproceedings{demszky2020goemotions,
  author    = {Demszky, Dorottya and Movshovitz-Attias, Dana and Ko, Jeongwoo and Cowen, Alan and Nemade, Gaurav and Ravi, Sujith},
  title     = {{GoEmotions}: A Dataset of Fine-Grained Emotions},
  booktitle = {Proceedings of the 58th Annual Meeting of the Association for Computational Linguistics},
  year      = {2020},
  pages     = {4040--4054},
  publisher = {Association for Computational Linguistics},
  url       = {https://aclanthology.org/2020.acl-main.372/}
}

@inproceedings{rashkin2019towards,
  author    = {Rashkin, Hannah and Smith, Eric Michael and Li, Margaret and Boureau, Y-Lan},
  title     = {Towards Empathetic Open-Domain Conversation Models: A New Benchmark and Dataset},
  booktitle = {Proceedings of the 57th Annual Meeting of the Association for Computational Linguistics},
  year      = {2019},
  pages     = {5370--5381},
  publisher = {Association for Computational Linguistics},
  url       = {https://aclanthology.org/P19-1534/}
}

@book{bowlby1969attachment,
  author    = {Bowlby, John},
  title     = {Attachment and Loss: Volume I: Attachment},
  publisher = {Basic Books},
  year      = {1969}
}

@article{mikulincer2003attachment,
  author  = {Mikulincer, Mario and Shaver, Phillip R. and Pereg, Daphna},
  title   = {Attachment Theory and Affect Regulation: The Dynamics, Development, and Cognitive Consequences of Attachment-Related Strategies},
  journal = {Motivation and Emotion},
  year    = {2003},
  volume  = {27},
  number  = {2},
  pages   = {77--102},
  doi     = {10.1023/A:1024515519160}
}

@incollection{waters2019attachment,
  author    = {Waters, Theodore E. A. and Waters, Harriet S.},
  title     = {Measuring Attachment Representations},
  editor    = {Cassidy, Jude and Shaver, Phillip R.},
  booktitle = {Handbook of Attachment: Theory, Research, and Clinical Applications},
  edition   = {3},
  publisher = {Guilford Press},
  year      = {2019},
  pages     = {235--260}
}

@book{ainsworth1978patterns,
  author    = {Ainsworth, Mary D. S. and Blehar, Mary C. and Waters, Everett and Wall, Sally N.},
  title     = {Patterns of Attachment: A Psychological Study of the Strange Situation},
  publisher = {Lawrence Erlbaum Associates},
  year      = {1978}
}

@article{rogers1957necessary,
  author  = {Rogers, Carl R.},
  title   = {The Necessary and Sufficient Conditions of Therapeutic Personality Change},
  journal = {Journal of Consulting Psychology},
  year    = {1957},
  volume  = {21},
  number  = {2},
  pages   = {95--103},
  doi     = {10.1037/h0045357}
}

@book{truax1967toward,
  author    = {Truax, Charles B. and Carkhuff, Robert R.},
  title     = {Toward Effective Counseling and Psychotherapy: Training and Practice},
  publisher = {Aldine},
  year      = {1967}
}

@book{stern1985interpersonal,
  author    = {Stern, Daniel N.},
  title     = {The Interpersonal World of the Infant},
  publisher = {Basic Books},
  year      = {1985}
}

@incollection{feeney2004transfer,
  author    = {Feeney, Brooke C. and Collins, Nancy L.},
  title     = {Interpersonal Safe Haven and Secure Base Caregiving Processes in Adulthood},
  editor    = {Rholes, W. Steven and Simpson, Jeffry A.},
  booktitle = {Adult Attachment: Theory, Research, and Clinical Implications},
  publisher = {Guilford Press},
  year      = {2004},
  pages     = {300--338}
}

@misc{cbsnews2024moderators,
  author  = {Stahl, Lesley and Chasan, Aliza and Bar-On, Shachar and Jung, Jinsol},
  title   = {Kenyan Workers with {AI} Jobs Thought They Had Tickets to the Future Until the Grim Reality Set In},
  year    = {2024},
  month   = nov,
  journal = {CBS News / 60 Minutes},
  url     = {https://www.cbsnews.com/news/ai-work-kenya-exploitation-60-minutes/},
  note    = {Accessed 2026-04-13}
}

@misc{sfgate2024scaleai,
  author  = {Council, Stephen},
  title   = {{SF} Tech Startup {Scale AI}, Worth \$13.8B, Accused of Widespread Wage Theft},
  year    = {2024},
  month   = dec,
  journal = {SFGATE},
  url     = {https://www.sfgate.com/tech/article/sf-tech-startup-scale-ai-sued-wage-theft-19976761.php},
  note    = {Accessed 2026-04-13}
}

@book{kish1965survey,
  author    = {Leslie Kish},
  title     = {Survey Sampling},
  year      = {1965},
  publisher = {John Wiley \& Sons},
  address   = {New York}
}

@article{thompson2012icc,
  author  = {Thompson, David M. and Fernald, Douglas H. and Mold, James W.},
  title   = {Intraclass Correlation Coefficients Typical of Cluster-Randomized Studies: Estimates From the Robert Wood Johnson Prescription for Health Projects},
  journal = {The Annals of Family Medicine},
  year    = {2012},
  volume  = {10},
  number  = {3},
  pages   = {235--240},
  doi     = {10.1370/afm.1347}
}

@article{davis2010interviewer,
  author  = {Davis, R. E. and Couper, M. P. and Janz, N. K. and Caldwell, C. H. and Resnicow, K.},
  title   = {Interviewer Effects in Public Health Surveys},
  journal = {Health Education Research},
  year    = {2010},
  volume  = {25},
  number  = {1},
  pages   = {14--26},
  doi     = {10.1093/her/cyp046}
}

@article{bradley1952rank,
  author  = {Bradley, Ralph Allan and Terry, Milton E.},
  title   = {Rank Analysis of Incomplete Block Designs: I. The Method of Paired Comparisons},
  journal = {Biometrika},
  volume  = {39},
  number  = {3/4},
  pages   = {324--345},
  year    = {1952},
  doi     = {10.2307/2334029}
}

@misc{rolnick2017deep,
  author       = {Rolnick, David and Veit, Andreas and Belongie, Serge and Shavit, Nir},
  title        = {Deep Learning Is Robust to Massive Label Noise},
  year         = {2017},
  eprint       = {1705.10694},
  archivePrefix= {arXiv},
  primaryClass = {cs.LG},
  url          = {https://arxiv.org/abs/1705.10694}
}

@inproceedings{natarajan2013learning,
  author    = {Natarajan, Nagarajan and Dhillon, Inderjit S. and Ravikumar, Pradeep K. and Tewari, Ambuj},
  title     = {Learning with Noisy Labels},
  booktitle = {Advances in Neural Information Processing Systems 26 (NeurIPS 2013)},
  year      = {2013},
  url       = {https://proceedings.neurips.cc/paper/2013/hash/3871bd64012152bfb53fdf04b401193f-Abstract.html}
}

@article{song2023noisy,
  author  = {Song, Hwanjun and Kim, Minseok and Park, Dongmin and Shin, Yooju and Lee, Jae-Gil},
  title   = {Learning from Noisy Labels with Deep Neural Networks: A Survey},
  journal = {IEEE Transactions on Neural Networks and Learning Systems},
  year    = {2023},
  volume  = {34},
  number  = {11},
  pages   = {8135--8153}
}

@book{hyland2005metadiscourse,
  author    = {Hyland, Ken},
  title     = {Metadiscourse: Exploring Interaction in Writing},
  publisher = {Continuum},
  year      = {2005}
}

@inproceedings{li2017dailydialog,
  author    = {Li, Yanran and Su, Hui and Shen, Xiaoyu and Li, Wenjie and Cao, Ziqiang and Niu, Shuzi},
  title     = {DailyDialog: A Manually Labelled Multi-turn Dialogue Dataset},
  booktitle = {Proceedings of the Eighth International Joint Conference on Natural Language Processing (Volume 1: Long Papers)},
  year      = {2017},
  pages     = {986--995},
  publisher = {Asian Federation of Natural Language Processing},
  url       = {https://aclanthology.org/I17-1099/}
}

@incollection{schwarz2011feelings,
  author    = {Schwarz, Norbert},
  title     = {Feelings-as-Information Theory},
  booktitle = {Handbook of Theories of Social Psychology},
  publisher = {SAGE Publications},
  year      = {2011},
  volume    = {1},
  pages     = {289--308}
}

@article{barsalou2008grounded,
  author  = {Barsalou, Lawrence W.},
  title   = {Grounded Cognition},
  journal = {Annual Review of Psychology},
  year    = {2008},
  volume  = {59},
  pages   = {617--645}
}

@inproceedings{quanhaase2024parasocial,
  author    = {Maeda, Takuya and Quan-Haase, Anabel},
  title     = {When Human-{AI} Interactions Become Parasocial},
  booktitle = {Proceedings of the 2024 ACM Conference on Fairness, Accountability, and Transparency},
  year      = {2024},
  pages     = {1914--1928},
  doi       = {10.1145/3630106.3659003}
}

@misc{fang2025psychosocial,
  author       = {Fang, Cathy Mengying and Pataranutaporn, Pat and Lampe, Michael and Phang, Jason and Agarwal, Sandhini and Gileadi, Nava and Wang, Xiaomeng and Maes, Pattie},
  title        = {How {AI} and Human Behaviors Shape Psychosocial Effects of Chatbot Use: A Longitudinal Randomized Controlled Study},
  year         = {2025},
  eprint       = {2503.17473},
  archivePrefix= {arXiv},
  primaryClass = {cs.HC},
  url          = {https://arxiv.org/abs/2503.17473}
}

@misc{phang2025affective,
  author       = {Phang, Jason and Lampe, Michael and Agarwal, Sandhini and Fang, Cathy Mengying and Pataranutaporn, Pat and Maes, Pattie},
  title        = {Investigating Affective Use and Emotional Well-Being on {ChatGPT}},
  year         = {2025},
  eprint       = {2504.03888},
  archivePrefix= {arXiv},
  primaryClass = {cs.HC},
  url          = {https://arxiv.org/abs/2504.03888}
}

@article{kim2008culture,
  author  = {Kim, Heejung S. and Sherman, David K. and Taylor, Shelley E.},
  title   = {Culture and Social Support},
  journal = {American Psychologist},
  year    = {2008},
  volume  = {63},
  number  = {6},
  pages   = {518--526}
}

@article{de2011emotion,
  author  = {De Leersnyder, Jozefien and Mesquita, Batja and Kim, Heejung S.},
  title   = {Where Do My Emotions Belong? A Study of Immigrants' Emotional Acculturation},
  journal = {Personality and Social Psychology Bulletin},
  year    = {2011},
  volume  = {37},
  number  = {4},
  pages   = {451--463},
  doi     = {10.1177/0146167211399103}
}

@article{elfenbein2002universality,
  author  = {Elfenbein, Hillary Anger and Ambady, Nalini},
  title   = {On the Universality and Cultural Specificity of Emotion Recognition: A Meta-Analysis},
  journal = {Psychological Bulletin},
  year    = {2002},
  volume  = {128},
  number  = {2},
  pages   = {203--235}
}

@article{henrich2010weirdest,
  author  = {Henrich, Joseph and Heine, Steven J. and Norenzayan, Ara},
  title   = {The Weirdest People in the World?},
  journal = {Behavioral and Brain Sciences},
  year    = {2010},
  volume  = {33},
  number  = {2--3},
  pages   = {61--83}
}

@inproceedings{GoharCheng2023IntersectionalFairness,
  author    = {Gohar, Usman and Cheng, Lu},
  title     = {A Survey on Intersectional Fairness in Machine Learning: Notions, Mitigation, and Challenges},
  booktitle = {Proceedings of the Thirty-Second International Joint Conference on Artificial Intelligence},
  year      = {2023},
  pages     = {6619--6627},
  note      = {Survey Track},
  url       = {https://www.ijcai.org/proceedings/2023/742}
}

@book{roberts2019behind,
  author    = {Roberts, Sarah T.},
  title     = {Behind the Screen: Content Moderation in the Shadows of Social Media},
  publisher = {Yale University Press},
  address   = {New Haven},
  year      = {2019}
}

@article{klie2024analyzing,
  title     = {Analyzing Dataset Annotation Quality Management in the Wild},
  author    = {Klie, Jan-Christoph and Eckart de Castilho, Richard and Gurevych, Iryna},
  journal   = {Computational Linguistics},
  volume    = {50},
  number    = {3},
  month     = sep,
  year      = {2024},
  address   = {Cambridge, MA},
  publisher = {MIT Press},
  url       = {https://aclanthology.org/2024.cl-3.1/},
  doi       = {10.1162/coli_a_00516},
  pages     = {817--866}
}

@inproceedings{huang2023worker,
  author    = {Huang, Olivia and Fleisig, Eve and Klein, Dan},
  title     = {Incorporating Worker Perspectives into {MT}urk Annotation Practices for {NLP}},
  booktitle = {Proceedings of the 2023 Conference on Empirical Methods in Natural Language Processing},
  year      = {2023},
  month     = dec,
  pages     = {1010--1028},
  address   = {Singapore},
  publisher = {Association for Computational Linguistics},
  doi       = {10.18653/v1/2023.emnlp-main.64},
  url       = {https://aclanthology.org/2023.emnlp-main.64/}
}

@article{sharma2023towards,
  title={Towards Understanding Sycophancy in Language Models},
  author={Sharma, Mrinank and Tong, Meg and Korbak, Tomasz and Duvenaud, David and Askell, Amanda and Bowman, Samuel R. and others},
  journal={arXiv preprint arXiv:2310.13548},
  year={2023}
}

@article{perez2022discovering,
  title={Discovering Language Model Behaviors with Model-Written Evaluations},
  author={Perez, Ethan and Ringer, Sam and Luko{\v{s}}i{\=u}t{\.e}, Kamil{\.e} and others},
  journal={arXiv preprint arXiv:2212.09251},
  year={2022}
}

@article{shapira2026rlhf,
  title={How RLHF Amplifies Sycophancy},
  author={Shapira, Itai and Benad{\`e}, Gerdus and Procaccia, Ariel D.},
  journal={arXiv preprint arXiv:2602.01002},
  year={2026}
}

@inproceedings{lambert2024tulu,
  title = {{T\"ulu 3}: Pushing Frontiers in Open Language Model Post-Training},
  author = {Lambert, Nathan and
            Morrison, Jacob and
            Pyatkin, Valentina and
            Huang, Shengyi and
            Ivison, Hamish and
            Brahman, Faeze and
            Miranda, Lester James V. and
            Liu, Alisa and
            Dziri, Nouha and
            Lyu, Xinxi and
            Gu, Yuling and
            Malik, Saumya and
            Graf, Victoria and
            Hwang, Jena D. and
            Yang, Jiangjiang and
            Le Bras, Ronan and
            Tafjord, Oyvind and
            Wilhelm, Chris and
            Soldaini, Luca and
            Smith, Noah A. and
            Wang, Yizhong and
            Dasigi, Pradeep and
            Hajishirzi, Hannaneh},
  booktitle = {Second Conference on Language Modeling},
  year = {2025},
  url = {https://openreview.net/forum?id=i1uGbfHHpH}
}

\end{document}